\documentclass[journal]{IEEEtran}

\usepackage{amssymb,amsmath,amsfonts}
\usepackage{algorithmic}
\usepackage{algorithm}
\usepackage{array}
\usepackage{bbold}
\usepackage{booktabs}
\usepackage{colortbl}
\usepackage[hidelinks]{hyperref}
\usepackage{graphicx}
\usepackage{makecell}
\usepackage[mathlines, switch]{lineno}
\usepackage{multirow}
\usepackage[sort&compress, numbers]{natbib}
\usepackage{pifont}
\usepackage{siunitx}
\usepackage{stfloats}
\usepackage[caption=false,font=footnotesize]{subfig}
\usepackage{textcomp}
\usepackage{xurl}
\usepackage{verbatim}
\usepackage[dvipsnames]{xcolor}

\hypersetup{
  colorlinks   = true, 
  urlcolor     = CadetBlue, 
  linkcolor    = Cerulean, 
  citecolor   = Maroon 
}

\begin{document}

\title{From Imitation to Exploration: End-to-end Autonomous Driving based on World Model}

\author{Yueyuan~Li, Mingyang~Jiang, Songan~Zhang, Jiale~Zhang, Chunxiang~Wang, and~Ming~Yang 
\thanks{This work is supported in part by the National Natural Science Foundation of China under Grant 62173228 \textit{(Corresponding author: Ming Yang, email: MingYANG@sjtu.edu.cn)}.}
\thanks{Yueyuan Li, Mingyang Jiang, Jiale Zhang, Chunxiang Wang, and Ming Yang are with the Department of Automation, Shanghai Jiao Tong University, Key Laboratory of System Control and Information Processing, Ministry of Education of China, Shanghai, 200240, CN (Yueyuan Li and Mingyang Jiang are co-first authors).}
\thanks{Songan Zhang is with the Global Institute of Future Technology, Shanghai Jiao Tong University, Shanghai, 200240, CN.}
}

\maketitle

\begin{abstract}

In recent years, end-to-end autonomous driving architectures have gained increasing attention due to their advantage in avoiding error accumulation. Most existing end-to-end autonomous driving methods are based on Imitation Learning (IL), which can quickly derive driving strategies by mimicking expert behaviors. However, IL often struggles to handle scenarios outside the training dataset, especially in high-dynamic and interaction-intensive traffic environments. In contrast, Reinforcement Learning (RL)-based driving models can optimize driving decisions through interaction with the environment, improving adaptability and robustness.

To leverage the strengths of both IL and RL, we propose RAMBLE, an end-to-end world model-based RL method for driving decision-making. RAMBLE extracts environmental context information from RGB images and LiDAR data through an asymmetrical variational autoencoder. A transformer-based architecture is then used to capture the dynamic transitions of traffic participants. Next, an actor-critic structure reinforcement learning algorithm is applied to derive driving strategies based on the latent features of the current state and dynamics. To accelerate policy convergence and ensure stable training, we introduce a training scheme that initializes the policy network using IL, and employs KL loss and soft update mechanisms to smoothly transition the model from IL to RL.

RAMBLE achieves state-of-the-art performance in route completion rate on the CARLA Leaderboard 1.0 and completes all 38 scenarios on the CARLA Leaderboard 2.0, demonstrating its effectiveness in handling complex and dynamic traffic scenarios. The model will be open-sourced upon paper acceptance at \url{https://github.com/SCP-CN-001/ramble} to support further research and development in autonomous driving.
\end{abstract}


\section{Introduction}

End-to-end autonomous driving systems have garnered increasing attention in recent years. This architecture directly learns driving decisions from sensor data, bypassing the complexity of multi-module designs and reducing the risk of error accumulation. An obstacle in establishing the end-to-end framework is handling the high-dimensional and multi-modal raw sensory inputs. World models can create implicit representations of the environment's current state and dynamics, providing an effective solution for processing perceptual information within end-to-end autonomous driving frameworks.

Existing end-to-end autonomous driving systems primarily rely on two learning paradigms: Imitation Learning (IL) and Reinforcement Learning (RL). In the IL paradigm, both observations and actions are arbitrary, allowing deep neural networks to capture complex features without significant concerns about stability in convergence \cite{hussein2017imitation}. This approach enables the rapid derivation of reasonable driving strategies \cite{wu2022trajectory, shao2023safety, jaeger2023hidden}. However, IL is heavily reliant on labeled data and struggles to generalize to unseen scenarios, particularly in complex and dynamic traffic environments \cite{lu2023imitation}. In contrast, the RL paradigm allows the agent to explore and refine its policy through continuous interaction with the environment. This enables the system to encounter a wider range of scenarios and better understand the causal relationships underlying the policy. To enhance the upper limit of capability to handle diverse and unpredictable scenarios, adapting the RL paradigm to autonomous driving is necessary \cite{wurman2022outracing}.

A significant challenge when leveraging the exploration mechanism in RL is the cold start problem. Exploration involves searching for the optimal policy within a vast action space, which typically requires substantial trial and error during the early stages of training. The agent must undergo thousands of steps of interaction with the environment to understand its dynamics and reward structure \cite{henderson2018deep}. Moreover, due to the nature of the Bellman equation, RL models must account for long-term rewards rather than focusing solely on immediate feedback \cite{ladosz2022exploration}. In the context of autonomous driving, the generation of simulated environmental data is relatively slow. These combined factors mean that training an RL-based driving model may take weeks to converge. To make the RL paradigm practical for autonomous driving, it is essential to accelerate and stabilize policy convergence.

This paper aims to address the aforementioned issues to enhance the generalization capability of autonomous driving systems across diverse scenarios. We propose RAMBLE\footnote{In this abbreviation, R, L, and A stand for the Reinforcement Learning Agent, MB is Model-based, and E is from End-to-end.}, an end-to-end world model-based RL method for driving decision-making. The contributions are as follows:

\begin{enumerate}
    \item We introduce an end-to-end, world model-based RL algorithm capable of processing multi-modal, high-dimensional sensory inputs that effectively addresses diverse and complex traffic environments.
    \item We propose a training strategy to address the cold start problem in RL by pretraining the world model, initializing the policy network with IL, and using KL loss and soft updates to smoothly transit the model to the RL training paradigm.
    \item We show that RAMBLE delivers state-of-the-art (SOTA) performance on the CARLA Leaderboard 1.0 and completes most of the highly interactive scenarios in CARLA Leaderboard 2.0, highlighting its ability to handle complex and dynamic traffic situations.
\end{enumerate}

\section{Related Works}

\subsection{End-to-end Driving with Deep Learning}

End-to-end driving methods directly learn driving policy from raw sensor data. This framework aims to avoid explicit feature extraction, which may cause a loss of crucial latent features and accumulate errors. Most end-to-end driving models fall into two main classes: IL and RL \cite{chib2024recent}.

In recent years, IL-based driving models have become mainstream in the field. They primarily focus on leveraging features from high-dimensional inputs and imitate expert behavior during policy learning for rapid convergence. This approach has been popular since the proposal of the first end-to-end driving model, DAVE-2 \cite{bojarski2016end}. Many IL-based driving models are focused on improving perception capabilities. Notable examples include Transfuser and InterFuser, which utilize transformer-based architectures to integrate features from multi-view images and LiDAR point clouds \cite{chitta2022transfuser, shao2023safety}. Models such as ReasonNet and CarLLaVA further refine performance by generating intermediate features with semantic meaning, improving context awareness \cite{shao2023reasonnet, renz2024carllava}. Some research attempts to ensemble outputs. For instance, TCP combines action commands with waypoint predictions, which becomes an effective improvement adopted by many subsequent works \cite{wu2022trajectory}. Similarly, Transfuser++ enhances performance by decoupling waypoint prediction from velocity estimation \cite{jaeger2023hidden}. However, the main drawback of IL-based driving models is their reliance on high-quality trajectory data or command records from experts \cite{xu2017end, chen2020learning, chen2022learning}, so their performance naturally degrades in corner cases or unseen scenarios.

While IL-based methods have thrived, RL-based driving models have encountered more challenges. Despite early efforts in specific scenarios \cite{sallab2017deep, zhang2019reinforcement, kendall2019learning}, researchers have found that training RL agents is costly due to the need for interaction with the environment. Additionally, model-free RL-based driving models face significant challenges in processing high-dimensional inputs and achieving stable convergence to a reasonable policy \cite{toromanoff2020end, zhao2022cadre}. While models like Roach and Think2Drive have demonstrated the potential of RL in driving decision-making, their reliance on privileged information makes them impractical for real-world applications \cite{zhang2021end, li2024think2drive}. To enable models to learn scenario context from complex raw sensor data, Peng \textit{et al.} proposed using expert driving behavior to guide the learning process, which has been shown to converge to safer policies \cite{peng2024learning}. Although RL-based models have a higher upper bound in the theoretical capability to handle diverse and complex traffic scenarios, they still lag behind IL-based models due to the challenges in implementation.

\subsection{World Model in Autonomous Driving}

The world model is a model-based RL algorithm introduced by Ha et al. in 2018 \cite{ha2018world}. The core idea of this method is to compress the observation at each time step using a variational autoencoder (VAE), followed by modeling the temporal dynamics with a recurrent neural network (RNN). Finally, a simple network is used to derive a behavior policy from the latent features extracted by the VAE and RNN. The goal of the World Model is to compress the high-dimensional, complex inputs in real-world tasks (such as visual observations) into lower-dimensional representations, enabling the RL-based policy network to perform behavior exploration more efficiently. Since the introduction of the Recurrent State-space Model (RSSM) in the Dreamer series \cite{hafner2023mastering}, the practical value of the world model has been significantly enhanced. RSSM adopts a more advanced state-space modeling approach, improving the model's adaptability in complex, dynamic environments.

The notable advantage of the world model lies in its ability to accurately model environmental dynamics and simulate the evolution of observations, which not only improves the sample efficiency of RL but also enhances data diversity for IL \cite{guan2024world}. This feature led to the first world model-based end-to-end autonomous driving method, MILE \cite{hu2022model}, followed by a series of studies utilizing the world model to optimize autonomous driving perception \cite{jia2023adriver, wang2024driving, min2023uniworld}, or generate synthetic data \cite{hu2023gaia, wang2024worlddreamer}.

Despite these advances, the structural complexity and resource demands, the world model has primarily been used to enhance IL-based methods in autonomous driving \cite{wang2024drivedreamer}. The field saw a breakthrough with Think2Drive \cite{li2024think2drive}, which successfully integrated a world model with an RL-based exploration mechanism for autonomous driving. However, Think2Drive relies on a significant amount of privileged information, such as HD maps, bounding boxes, and traffic states, which introduces a risk of error accumulation. Furthermore, the method requires training the RL model from scratch, which incurs a high time cost.

\section{Method}

\subsection{Overview}

\begin{figure*}
    \centering
    \includegraphics[width=\linewidth]{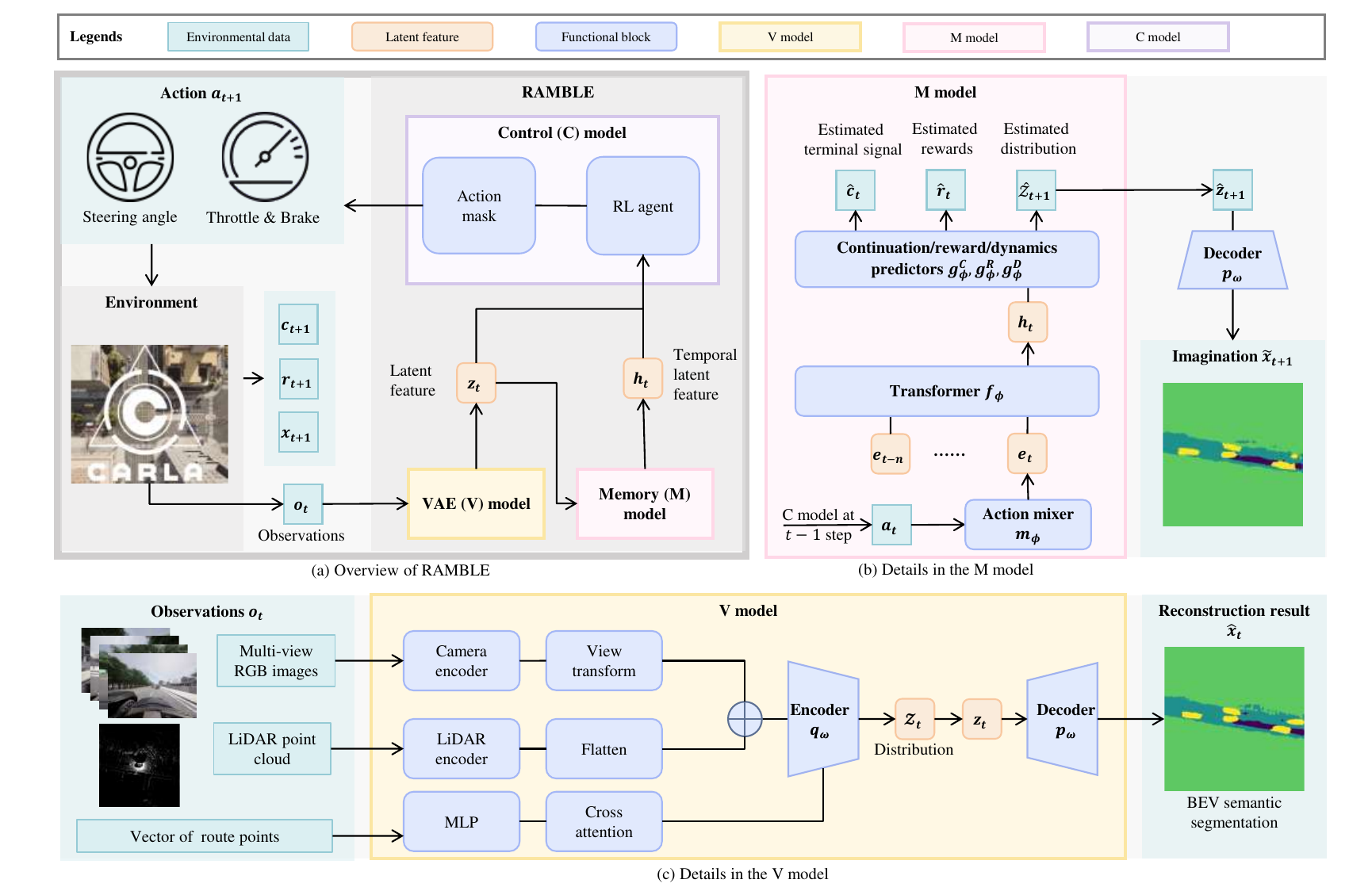}
    \caption{The overall structure of RAMBLE. The V model compresses multi-view RGB images, LiDAR point clouds, and route points to the latent feature $z_t$, which describes the current state. The M model takes the latent features $z_{t-n:t}$ and actions $a_{t-n:t}$ to predict the environmental dynamics by the latent feature $h_t$. The C model generates action commands based on $z_t$ and $h_t$.}
    \label{figure: RAMBLE}
\end{figure*}

The overall architecture of RAMBLE is illustrated in Figure \ref{figure: RAMBLE}. Our approach builds upon the framework introduced in the original world model \cite{ha2018world}. For the sake of clarity, we retain the naming conventions for the functional networks from the prior work. Specifically, the VAE (V) model refers to the network that compresses the current observation into a latent feature, the Memory (M) model extracts temporal dynamics from a sequence of observations, and the Control (C) model is responsible for decision-making based on the latent features.

We introduce an asymmetric VAE tailored for the high-dimensional multi-modal input in the V model (Section \ref{section: V-model}). In the M model, we adopt a transformer to extract temporal information (Section \ref{section: M-model}). The rewards in the C model are designed for the autonomous driving task (Section \ref{section: C-Model}). We outline our training procedure to enhance and stabilize the policy convergence process in Section \ref{section: training-recipe}. Finally, to improve the safety performance of RAMBLE, we implement an action mask for the control command (Section \ref{section: Action-mask}).

\subsection{V Model} \label{section: V-model}


The original V model was designed to process the $210\times160$ pixel images from the Atari environment \cite{ha2018world}. However, in the context of autonomous driving, the observation space is more complex for capturing the necessary information. Our input observations consist of multi-view RGB images and LiDAR point clouds. The RGB images are captured at a resolution of $640\times480$ from the front, left, right, and rear perspectives. The LiDAR data includes approximately 31,000 points from a 360$^\circ$ scan. To accommodate this, we adapt the V model to an asymmetric structure, where the inputs consist of multi-modal raw sensor data, and the output is a bird's-eye-view (BEV) semantic segmentation image.

The architecture preceding the asymmetric VAE is inspired by BEVFusion \cite{liang2022bevfusion}. Raw camera data is first processed by a Swin Transformer to extract multi-scale features \cite{liu2021swin}, which are then integrated into a feature map using a Feature Pyramid Network \cite{lin2017feature}. The image features are projected into the BEV and mapped onto 3D grids. Concurrently, LiDAR point clouds are voxelized and mapped to 3D grids using a PointPillars network \cite{lang2019pointpillars}. Since the point cloud features are inherently in BEV format, no additional view transformation is required. The image and point cloud features are then concatenated and fused through a Convolutional Neural Network (CNN). Finally, the route points are processed by a Multilayer Perceptron (MLP), which is integrated with the sensory data via a cross-attention mechanism before being passed to the VAE.

The feature extraction network and the VAE's encoder together are referred to as the observation encoder $q_\omega$, while the VAE's decoder is denoted as the observation decoder $p_\omega$. $\omega$ represents the parameters of the V model's network. To enhance the generalization ability of the whole model, the latent encoder does not directly output the latent feature $z_t$. Instead, it models the environment dynamics with a distribution $\mathcal{Z}_t$, from which the latent feature is sampled \cite{micheli2022transformers}.
\begin{align*}
    & \textrm{Observation encoder:} & z_t & \sim q_\omega(z_t|o_t) = \mathcal{Z}_t \\
    & \textrm{Observation decoder:} & \hat{x}_t & = p_\omega(z_t)
\end{align*}
here $o_t$ represents the raw sensor observations at time $t$, and $\hat{x}_t$ is the reconstructed observation. We use two different letters because the encoder's output is different from the input.

The encoder is heavy in parameters, so it is irrational to expect it to learn the necessary latent features by training directly alongside the rest of RAMBLE. To address this, we pre-train the encoder with BEV semantic segmentation images $x_t$ as the label. A detailed explanation of the training process can be found in Section \ref{section: training-recipe}. Given the imbalance in the semantic labels, we use sigmoid focal loss to handle the reconstruction loss \cite{lin2017focal}. The component loss functions used during the pre-training phase are defined as follows:
\begin{align*}
    \mathcal{P}_t & = x_t \cdot \sigma (\hat{x}_t) + (\mathbb{1} - x_t) \cdot (\mathbb{1} - \sigma(\hat{x}_t)) \\
    \mathcal{L}^\textrm{rec}_t(\omega) & = -\alpha(1-\mathcal{P}_t)^\gamma \cdot \log(\mathcal{P}_t) \\
    \mathcal{L}^\textrm{kl}_t(\omega) & = \textrm{KL}(q_\omega(z_t|o_t) \Vert p_\omega(z_t))
\end{align*}
where $\sigma(\cdot)$ means the sigmoid function. $\mathcal{P}^t$ is the probability of the reconstructed observation being correct. $\mathcal{L}^\textrm{rec}_t(\omega)$ is the reconstruction loss, whose class balancing factor is $\alpha$ and focusing parameter is $\gamma$. $\mathcal{L}^\textrm{kl}_t(\omega)$ is the KL divergence loss.

Gathering them together, we can obtain the loss function $\mathcal{L}^V_t(\omega)$ for pre-training the V model:
\begin{align}\label{equation: V-model-loss}
    \mathcal{L}^V_t(\omega)  = \mathcal{L}^\textrm{rec}_t (\omega) + k^\textrm{kl} \mathcal{L}^\textrm{kl}_t (\omega)
\end{align}
where $k^\textrm{kl}$ is a scalar factor that dynamically adjusts the weight of the KL divergence loss.

\subsection{M Model} \label{section: M-model}

The target of the M model is to capture temporal information from the sequence of latent features $\{z_{t-n}, \dots, z_t\}$. The initial version of the world model uses a Long Short-Term Memory (LSTM) network \cite{ha2018world}. Later, the transformer architecture has been proven to be more effective at capturing long-range dependencies in sequential data \cite{micheli2022transformers, zhang2024storm}. Therefore, we implement the M model with a transformer.

As is demonstrated in the following equations, the latent feature $z_t$ sampled from $\mathcal{Z}_t$ is combined with the action $a_t$, using MLP and concatenation. This manipulation is denoted as $m_\phi$, and the mixed feature is $e_t$. We then adopt a transformer $f_\phi$ with a stochastic attention mechanism to process the mixed feature. The output of the transformer is the latent feature $h_t$. Finally, $h_t$ is fed into three different MLPs: $g_\phi^D$ predicts the next environment dynamics distribution $\hat{\mathcal{Z}}_{t+1}$; $g_\phi^R$ predicts the current reward $\hat{r}_{t}$; and $g_\phi^C$ predicts the current traffic status $\hat{c}_t$, which decides the episode will continue or not. $\phi$ represents the parameters of the M model.
\begin{align*}
    &\textrm{Action mixer:} & e_t & = m_\phi(z_t, a_t) \\
    &\textrm{Sequence model:} & h_{1:T} & = f_\phi(e_{1:T})\\
    &\textrm{Dynamics predictor:} & \hat{z}_{t+1} & \sim g_\phi^D (\hat{z}_{t+1} | h_t) = \hat{\mathcal{Z}}_{t+1} \\
    &\textrm{Reward predictor:} & \hat{r}_t & = g_\phi^R (h_t) \\
    &\textrm{Continuation predictor:} & \hat{c}_t & = g_\phi^C (h_t)
\end{align*}
The M model's loss function comprises multiple components, each designed to optimize a specific objective. The dynamics loss, $\mathcal{L}^\textrm{dyn}_t$, and the representation loss, $\mathcal{L}^\textrm{rep}$, help predict the next distribution $\hat{Z}_{t+1}$ while ensuring that $h_t$ closely couples with $z_t$. The continuation prediction error is captured by $\mathcal{L}^\textrm{con}_t$, which utilizes binary cross-entropy loss.
\begin{align*}
    \mathcal{L}^\textrm{dyn}_t (\phi) & = \max(1, \textrm{KL}[\textrm{sg}(q_\phi(z_{t+1}|o_{t+1})) \Vert g_\phi^D(\hat{z}_{t+1}|h_t)]) \\
    \mathcal{L}^\textrm{rep}_t (\phi) & = \max(1, \textrm{KL}[q_\phi(z_{t+1}|o_{t+1}) \Vert \textrm{sg}(g_\phi^D(\hat{z}_{t+1}|h_t))]) \\
    \mathcal{L}^\textrm{con}_t (\phi) & = c_t \log \hat{c}_t + (1 - c_t) \log (1 - \hat{c}_t)
\end{align*}
Here, $\textrm{sg}(\cdot)$ denotes the stop-gradient function. By tuning the scales of $\mathcal{L}^\textrm{dyn}_t$ and $\mathcal{L}^\textrm{rep}_t$, we can regulate the magnitude and proportion of the gradient updates of V and M models.

Because of the complexity of the traffic scenario, the predicted reward is no longer represented by a single value. Instead, we implement multiple reward functions to evaluate driving performance from various perspectives (see Section \ref{section: C-Model}). To assist the M model in capturing the underlying dynamics of the environment, we guide it to predict the physical variables that are used to compute these rewards. Consequently, we modify the loss function to the Huber Loss. Both the predicted rewards and the ground truth are represented as vectors and normalized to the range $[-1, 1]$ before computing the loss.
\begin{align*}
    \mathcal{L}^\textrm{rew}_t (\phi) = \Vert \textrm{norm}(\hat{r}_t) - \textrm{norm}(r_t) \Vert_\textrm{Huber}
\end{align*}

Due to the heterogeneity in the modality of the input data and the prediction result of the dynamic predictor, the reconstruction loss, which is the most critical component of M model's loss, is computed by first decoding the estimated latent feature, $\hat{z}_{t+1}$, sampled from $\hat{\mathcal{Z}}_{t+1}$, to $\tilde{x}_{t+1}$ using the VAE decoder. This result is then compared with the ground truth observation, $x_{t+1}$, using the sigmoid focal loss.
\begin{align*}
    \tilde{x}_{t+1} & = p_\omega (\hat{z}_{t+1}) \\
    \tilde{\mathcal{P}}_{t+1} & = x_{t+1} \cdot \sigma (\tilde{x}_{t+1}) + (1 - x_{t+1}) \cdot (1 - \sigma(\tilde{x}_{t+1})) \\
    \mathcal{L}^\textrm{rec}_t (\phi) & = -\alpha(1-\tilde{\mathcal{P}}_{t+1})^\gamma \cdot \log(\tilde{\mathcal{P}}_{t+1})
\end{align*}

The total loss $\mathcal{L}^M_t$ is a weighted sum of the above:
\begin{align}\label{equation: M-model-loss}
    \mathcal{L}^M_t(\phi) =& \mathcal{L}^\textrm{rec}_t (\phi) + k^\textrm{dyn}\mathcal{L}^\textrm{dyn}_t(\phi) + k^\textrm{rep}\mathcal{L}^\textrm{rep}_t(\phi) \\ \notag
    &+ k^\textrm{rew}\mathcal{L}^\textrm{rew}_t(\phi) + k^\textrm{con}\mathcal{L}^\textrm{con}_t(\phi)
\end{align}
where $k^\textrm{dyn}$, $k^\textrm{rep}$, $k^\textrm{rew}$, and $k^\textrm{con}$ are the weight factors.

\subsection{C Model} \label{section: C-Model}

\subsubsection{RL Agent}

In this work, we adopt the Soft Actor-Critic (SAC) algorithm as the backbone \cite{haarnoja2018soft}, primarily due to its training stability and relative insensitivity to the design of reward functions. The RL agent outputs a waypoint relative to the ego vehicle in a polar coordinate system. This waypoint is then converted into steering, throttle, and brake commands by a PID controller. This approach eliminates the need to predict acceleration directly, the second derivative of displacement, thus simplifying the policy learning process \cite{jiang2025hope}.

\subsection{Step Reward}

\subsubsection{Reward for speed $r_\textrm{speed}$}

This reward encourages the vehicle to maintain an appropriate speed. The expression of $r_\textrm{speed}$ depends on $v_\textrm{lon}$, the velocity along the route. The vehicle receives a reward equal to that for reaching the maximum velocity, $v_{\max}$, to encourage stopping when ``should stop'' conditions, derived from prior knowledge, are met.
\begin{equation*}
    r_\textrm{speed} =
\begin{cases}
    min(v_\textrm{lon}, v_{\max}) & \textrm{if not ``should stop''} \\
    v_{\max} & \textrm{if ``should stop'' and } v=0
\end{cases}
\end{equation*}

\subsubsection{Reward for traveled distance $r_\textrm{distance}$}

The traveled distance is monitored at each step. Once the agent covers an additional meter, it receives a constant reward. This reward helps address the issue of reward sparsity, which can occur if the agent is only rewarded for its traveled distance at the end of an episode.

\subsubsection{Penalty for route deviation angle $r_\textrm{dev angle}$}

This penalty measures the angular difference $\theta_\textrm{diff}$ between the ego vehicle's heading and the route direction. The weight of this penalty will decrease if the ego vehicle has been stuck for a long time to encourage lane-changing behavior.
\begin{equation*}
r_\textrm{dev angle} =
\begin{cases}
    0 & \textrm{if } |\theta_\textrm{diff}| < \pi/6 \\
    -|\theta_\textrm{diff}| & \textrm{if } |\theta_\textrm{diff}| \geq \pi/6
\end{cases}
\end{equation*}

\subsubsection{Penalty for route distance deviation $r_\textrm{dev distance}$}

This penalty discourages the agent from deviating significantly from the route. It is calculated based on the lateral distance $d_\textrm{lat}$ between the ego vehicle and the center of the route. The weight of this penalty will decrease if the ego vehicle has been stuck for a long time.
\begin{equation*}
    r_\textrm{dev distance} = -{d_\textrm{lat}}^2
\end{equation*}

The total step reward $r_\textrm{step}$ is a weighted sum of all the rewards mentioned above.
\begin{equation}\label{equation: step-reward}
r_\textrm{step}=\max(r_\textrm{speed}+r_\textrm{distance}, v_{\max})+k_1\cdot r_\textrm{dev angle} + k_2\cdot r_\textrm{dev distance}
\end{equation}

\subsection{Terminal Status}

During the training process, to expedite the trial-and-error learning towards the correct behavior, we introduce certain terminal states that are easier to trigger. Rewards and penalties are assigned to these states.

\subsubsection{ Collision and penalty for collision $r_\textrm{collide}$}

An episode terminates if a collision with either dynamic or static obstacles is detected. The penalty is calculated based on the ego vehicle's velocity $v$, whose unit is m/s.
\begin{equation*}
    r_\textrm{collide} = -1 - v
\end{equation*}

\subsubsection{ Off-lane and penalty for off-lane $r_\textrm{off-lane}$}

An episode terminates if the lateral distance $d_\textrm{lat}$ exceeds 3.5 m. The penalty is given by:
\begin{equation*}
    r_\textrm{off-lane} = \textrm{clip}(\vert d_\textrm{lat} \vert \times 5, 20, 25)
\end{equation*}

\subsubsection{Timeout and penalty for timeout $r_\textrm{timeout}$}

An episode terminates if the vehicle takes too long to make progress. The penalty for a timeout discourages prolonged inactivity.
\begin{equation*}
    r_\textrm{timeout} = -10
\end{equation*}

These terminal conditions are detected independently, allowing them to be triggered simultaneously. 

Once an episode terminates, the route completion reward $r_\textrm{complete}$ is calculated based on both the traveled distance $d$ and the route completion rate $RC$. Since the route completion rate is proportional, and the route length can vary from 200 to 4000 m, we use the traveled distance to compensate for this variation.
\begin{equation*}
    r_\textrm{complete} = d / 100 + RC/4
\end{equation*}

The total terminal reward $r_\textrm{term}$ is the sum of all the rewards mentioned above.
\begin{equation}\label{equation: terminal-reward}
    r_\textrm{term} = r_\textrm{collide} + r_\textrm{off-lane} + r_\textrm{timeout} + r_\textrm{complete}
\end{equation}

\subsection{Training Recipe} \label{section: training-recipe}

The complexity of RAMBLE makes it challenging to simultaneously train V, M, and C models from scratch. To improve both the efficiency and stability of the training process, we propose the following sequential training strategy.

\subsubsection{Pre-train the V and M model}

We first collect approximately $10^5$ frames of data under diverse weather and lighting conditions from a variety of traffic scenarios to construct an offline dataset. The V model is first solely pre-trained using the loss function in Equation \ref{equation: V-model-loss}, where we control the scalar factor $k^\textrm{kl}$ contributing between 5\% and 10\% of the total loss. Then we freeze the V model and update the M model using the loss function in Equation \ref{equation: M-model-loss} before unfreezing the V model for jointly training. During training, we increase the weight factors $k^\textrm{dyn}$, $k^\textrm{rep}$, $k^\textrm{rew}$, and $k^\textrm{con}$ gradually to allow for more emphasis on the dynamics-related aspects of the task, with contribution to Equation \ref{equation: M-model-loss} of 5\%-10\%, 1\%–5\%, 5\%–10\%, and 5\%–10\%, respectively. At this stage, the role of the V model shifts from simply reconstructing a BEV semantic image to generating a hidden feature that encapsulates as much useful information as possible for dynamic prediction.

\subsubsection{Pre-train RAMBLE with IL}

With the pre-trained V and M models providing inputs, the networks in the C model will be initialized by imitating the expert demonstration. The imitation learning loss function, $\mathcal{L}^\textrm{IL}$, is calculated based on the distance between the C model's output waypoint and the expert's output waypoint, $d_\textrm{waypoint}$, as well as the heading difference  $\theta_\textrm{waypoint}$ computed from the center of the ego vehicle to the two waypoints.

\begin{equation}\label{equation: C-model-IL-loss}
    \mathcal{L}^\textrm{IL} = d_\textrm{waypoint} + 10\cdot \theta_\textrm{waypoint}
\end{equation}

In addition to improving the robustness of the behavior cloning model, we add a uniform random noise of about 0.2 to the expert steer and throttle.

\subsubsection{Train RAMBLE online with RL}

While pre-training policy using IL was applied for RL tasks in previous studies, the uninitialized value function and inherent difficulty in balancing exploration and exploitation can easily cause the pre-training to lose its effect \cite{ramrakhya2023pirlnav}. In fact, we have observed a similar phenomenon that simply initializing RL with pre-trained parameters would result in a sudden drop in performance, and the agent needs to learn from scratch again. To deal with it, we mainly apply two methods in our work:
\begin{itemize}
    \item We apply the learning rate schedule proposed in \cite{ramrakhya2023pirlnav}, where the learning rate scheduling is divided into 3 phases. In the second phase, we gradually increase the actor learning rate and decrease the critic learning rate, and in the first and the last phases, they remain unchanged. This approach makes a smooth transition when the critic network is not fully trained.
    \item We constrain the update of the actor parameters using KL divergence between the action distribution of the actor output and the expert imitator \cite{nair2020awac}, which helps to prevent irrational exploration from the agent.
\end{itemize}

In practice, we use pre-trained parameters to initialize two actor networks, the primary network $\pi_{\theta_p}$, which outputs the action to interact with the environment,
and the anchor network $\pi_{\theta_a}$, which provides a reference anchoring action to calculate the KL constraint. The KL loss is added to the original RL loss on the actor as:
\begin{equation}\label{equation: rl-with-kl}
    \mathcal{L}^\textrm{RL}_t(\theta_p) = \mathcal{L}^{SAC}(\theta_p) + \textrm{KL}[\pi_{\theta_p}(\cdot|h_t) || \pi_{\theta_a}(\cdot|h_t)],
\end{equation}
where $L^{SAC}(\theta_p)$ is the original loss function on the actor network from SAC. Moreover, to guarantee that the modified loss function does not influence the final convergence of RL training, we apply the soft update with a soft update coefficient $\tau$ on the anchor network $\pi_{\theta_a}$ as:
\begin{equation}\label{equation: soft-update-anchor}
    \pi_{\theta_a} = \tau\cdot\pi_{\theta_p} + (1-\tau)\cdot\pi_{\theta_a}.
\end{equation}
It is straightforward to validate that the modified loss function in Equation \ref{equation: rl-with-kl} shares the same convergent policy as the original loss function. For any convergent policy $\pi_{\theta^*}$ of $\mathcal{L}_{SAC}(\theta_p)$, the soft update in Equation \ref{equation: soft-update-anchor} leads to $\theta_p = \theta_a = \theta^*$, making $\text{KL}[\pi_{\theta_p}(\cdot|h_t) || \pi_{\theta_a}(\cdot|h_t)]=0$ and $\mathcal{L}^\textrm{RL}_t(\theta_p)$ reaching its minimium at $\mathcal{L}^{SAC}(\theta_p)$. This indicates that while IL pre-training helps the RL training for stable update, it does not necessarily constrain the affordance of the RL policy.

\begin{table*}[htb]

    \centering
    \caption{Performance comparison in CARLA Leaderboard 1.0}
    \label{table: performance-L10}
    \setlength\tabcolsep{4pt}

    {\renewcommand{\arraystretch}{1.3}
    \begin{tabular}{l|ccc|ccccccccc} \toprule[2pt]
    \textbf{Method} & \textbf{RC} & \textbf{DS} & \textbf{IS} & \makecell{\textbf{Collisions}\\\textbf{pedestrians}} & \makecell{\textbf{Collisions}\\\textbf{vehicles}} & \makecell{\textbf{Collisions}\\\textbf{layout}} & \makecell{\textbf{Red light}} & \makecell{\textbf{Stop sign}} & \makecell{\textbf{Off-road}} & \makecell{\textbf{Agent}\\\textbf{blocked}} & \makecell{\textbf{Route}\\\textbf{deviations}} & \makecell{\textbf{Route}\\\textbf{timeout}} \\ \hline
    & \%, $\uparrow$ & \%, $\uparrow$ & $\uparrow$ & \#/km, $\downarrow$ & \#/km, $\downarrow$ & \#/km, $\downarrow$ & \#/km, $\downarrow$ & \#/km, $\downarrow$ & \#/km, $\downarrow$ & \#/km, $\downarrow$ & \#/km, $\downarrow$ & \#/km, $\downarrow$ \\ \hline
    Roach-RL \cite{zhang2021end} & 96.9 & 87.3 & 0.90 & 0.00 &  0.16 & 0.01 & 0.00 & 0.02 & 0.00 & 0.11 & 0.00 & 0.03 \\
    Think2Drive$^*$ \cite{li2024think2drive} & 99.7 & 90.2 & - & - & - & - & - & - & - & - & - & - \\ \hline
    ReasonNet$^*$ \cite{shao2023reasonnet} & 89.9 & \textbf{80.0} & \textbf{0.89} & 0.02 & 0.13 & 0.01 & 0.08 & 0.00 & 0.04 & 0.33 & 0.00 & 0.01 \\
    LAV \cite{chen2022learning} & 84.2 & 54.4 & 0.67 & 0.00 & 0.27 & 0.16 & 0.04 & 0.62 & 0.05 & 0.12 & 0.32 & 0.14 \\
    TransFuser++ \cite{jaeger2023hidden} & 97.2 & 70.9 & 0.73 & 0.00 & 0.24 & 0.15 & 0.05 & 0.36 & 0.07 & 0.09 & 0.00 & 0.06 \\
    TCP \cite{wu2022trajectory} & 88.6 & 52.8 & 0.58 & 0.00 & 0.44 & 0.84 & 0.07 & 0.50 & 0.17 & 0.62 & 0.00 & 0.07 \\ \hline
    RAMBLE (Ours) & \textbf{98.2} & 39.4 & 0.40 & 0.04 & 0.37 & 0.02 & 1.76 & 0.35 & 0.06 & 0.02 & 0.00 & 0.00 \\ \bottomrule[2pt]
    \end{tabular}
    }

    \vspace{0.1cm}
    \scriptsize
    \textit{Notes}:
    Roach-RL and Think2Drive are expert models having access to privileged information from the simulator's backend. The methods marked with $*$ are not open-source, and the testing reports are sourced from their original papers.
\end{table*}

\subsection{Action Mask} \label{section: Action-mask}

RL has long faced efficiency and safety challenges during the training and deployment stages. Action masking serves as a practical technique to filter out invalid actions, thereby improving both exploration efficiency and operational safety \cite{huang2020closer}. In the context of autonomous driving, when the agent explores using a randomly sampled action from the output distribution, there is a high probability that it will reach a state where it collides with obstacles, causing the vehicle to become stuck and unable to proceed.

To address this issue, we adopt an action mask based on the point cloud data, following the method proposed in our previous work \cite{jiang2025hope}. The action mask functions as a differentiable post-processing module that clips the output trajectories to ensure they are collision-free concerning obstacles detected in the current frame. Although only static obstacles are considered, the action mask imposes a fundamental constraint, preventing collisions with static objects, while leaving room for the agent to learn interactions with dynamic entities.


\section{Experiment and Results}

\begin{table*}[htb]
    \centering
    \caption{Performance of RAMBLE in the 38 scenarios intercepted from CARLA Leaderboard 2.0}
    \label{table: performance-scenarios-L20}

    {\renewcommand{\arraystretch}{1.6}
    \begin{tabular}{lccc|lccc} \toprule[2pt]
    \textbf{Scenario} & \textbf{RC (\%)} &  \textbf{IP} & \textbf{DS (\%)}
    & \textbf{Scenario} & \textbf{RC (\%)} &  \textbf{IP} & \textbf{DS (\%)}  \\
    \hline
    Accident & 72.73 & 0.86 & 61.59
    & \makecell[l]{Accident\\TwoWays} & 82.29 & 0.89 & 73.60\\ \hline
    \makecell[l]{Construction\\Obstacle} & 76.10 & 0.89 & 68.14
    & \makecell[l]{Construction\\ObstacleTwoWays} & 83.28 & 0.85 & 71.15\\ \hline
    \makecell[l]{Blocked\\Intersection} & 86.20 & 0.56 & 44.52
    & ControlLoss & 95.84 & 0.68 & 64.47 \\ \hline
    \makecell[l]{Crossing\\BicycleFlow} & 97.78 & 0.76 & 74.40
    & \makecell[l]{DynamicObject\\Crossing} & 81.45 & 0.66 & 54.95\\ \hline
    EnterActorFlow & 99.99 & 0.80 & 80.11
    & EnterActorFlowV2 & 99.99 & 0.96 & 96.51\\ \hline
    HazardAtSideLane & 99.99 & 0.42 & 41.67
    & \makecell[l]{HazardAtSideLane\\TwoWays} & 92.62 & 0.92 & 86.26 \\ \hline
    HardBreakRoute & 96.39 & 0.60 & 56.92
    & HighwayCutIn & 97.60 & 0.93 & 91.03 \\ \hline
    HighwayExit & 99.99 & 0.57 & 57.04
    & InvadingTurn & 91.53 & 0.64 & 55.94\\ \hline
    \makecell[l]{Interurban\\ActorFlow} & 99.99 & 0.91 & 90.85
    & \makecell[l]{InterurbanAdvanced\\ActorFlow} & 99.99 & 0.57  & 57.02\\ \hline
    \makecell[l]{MergerInto\\SlowTraffic} & 99.99 & 0.75 & 75.30
    & \makecell[l]{MergerInto\\SlowTrafficV2} & 98.20 & 0.54 & 53.73\\ \hline
    \makecell[l]{NonSignalized\\JunctionLeftTurn} & 97.83 & 0.50 & 48.81
    & \makecell[l]{NonSignalized\\JunctionRightTurn} & 97.37 & 0.56 & 54.92\\ \hline
    \makecell[l]{OppositeVehicle\\RunningRedLight} & 98.02 & 0.84 & 83.36
    & \makecell[l]{OppositeVehicle\\TakingPriority} & 97.44 & 0.79 & 77.15\\ \hline
    ParkedObstacle & 68.27 & 0.89 & 60.63
    & \makecell[l]{ParkedObstacle\\TwoWays} & 71.39 & 0.92 & 64.65\\ \hline
    \makecell[l]{ParkingCrossing\\Pedestrian} & 94.52 & 0.74 & 72.63
    & ParkingExit & 95.84 & 0.60 & 57.34\\ \hline
    StaticCutIn & 99.99 & 0.88 & 97.81
    & ParkingCutIn & 90.16 & 0.81 & 74.68\\ \hline
    \makecell[l]{Pedestrian\\Crossing} & 99.32 & 0.69 & 68.26
    & PriorityAtJunction & 99.99 & 0.81 & 81.16\\ \hline
    \makecell[l]{SignalizedJunction\\LeftTurn} & 94.56 & 0.56 & 53.90
    & \makecell[l]{SignalizedJunction\\RightTurn} & 99.51 & 0.63 & 63.51\\ \hline
    \makecell[l]{VehicleTurning\\Route} & 96.43 & 0.66 & 64.11
    & \makecell[l]{VehicleTurning\\RoutePedestrian} & 98.36 & 0.64 & 63.58\\ \hline
    \makecell[l]{VehicleOpensDoor\\TwoWays} & 94.61 & 0.51 & 48.55
    & \makecell[l]{YieldTo\\EmergencyVehicle} & 99.99 & 0.64 & 64.36 \\
    \bottomrule[2pt]
    \end{tabular}
    }

\end{table*}

\subsection{Device}

We use a server equipped with two AMD EPYC 7713 CPUs (256 cores), 128 GB of memory, and one NVIDIA RTX 4090 GPU for training and testing our method.

\subsection{Performance on CARLA Leaderboard 1.0}

We selected the CARLA Leaderboard 1.0 as the benchmark for evaluating our autonomous driving model, as it is the most widely recognized and established platform for comparing autonomous driving systems \cite{chib2024recent}. The validation set of the leaderboard includes 26 routes across three different maps, with route lengths ranging from 800 to 4400m. The primary tasks in Leaderboard 1.0 focus on lane-keeping and navigating junctions, while specialized tasks such as emergency avoidance, merging, and negotiation are not included in this benchmark. Testing on Leaderboard 1.0 demonstrates our model’s stability in long-term lane keeping.

For comparison, we selected representative SOTA methods \cite{zhang2021end, jia2023think, shao2023reasonnet, chen2022learning, jaeger2023hidden, wu2022trajectory}. The coach RL model for Roach represents model-free reinforcement learning (RL) with privileged information \cite{zhang2021end}, while Think2Drive exemplifies model-based RL with privileged information \cite{li2024think2drive}. When sufficient information is provided, these two methods represent the best performance achieved in prior RL-based approaches. Additionally, LAV, ReasonNet, Transfuser++, and TCP are IL-based methods, each following distinct design principles, which have also demonstrated strong performance on Leaderboard 1.0.

In Table \ref{table: performance-L10}, we present a performance comparison between the baselines and RAMBLE, using the evaluation metrics defined by CARLA. To ensure fairness, we reproduced the performance of all open-source methods on our device, testing each across all routes three times with different random seeds. Due to the unavailability of the source code for Think2Drive and ReasonNet, we refer to their self-reported results.

The route completion rate (RC) of RAMBLE exceeds that of both IL methods and RL experts, highlighting that with appropriate compression of input data by the V and M models, RL-based approaches are well-suited for long-term driving decision-making.

The driving score (DS), however, is relatively low due to frequent infraction penalty (IP) related to red lights and stop signs. This is attributed to the fact that, due to limited human resources, we did not implement a specific penalty mechanism for such infractions. Designing and enforcing these penalties requires considerable manual effort to define and apply traffic rules. While ensuring legal compliance is important, it does not directly contribute to the core innovations or performance improvements of our method. As such, we have prioritized the primary objectives of the algorithm and plan to address this engineering challenge in future work.

\begin{figure*}[!p]
    \centering
    \includegraphics[width=\linewidth]{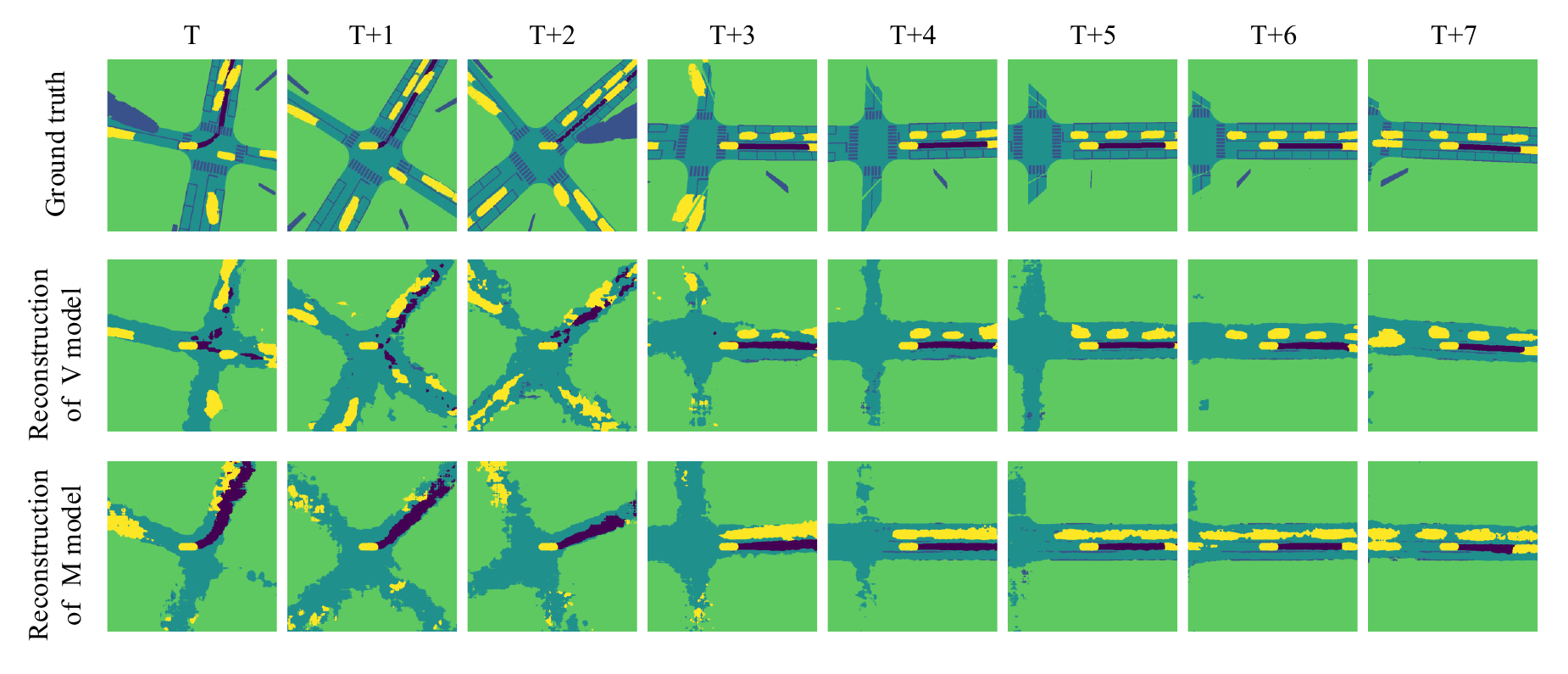}
    \caption{A visualization of the V model's state latent features and the M model's estimated state latent feature.}
    \label{figure: V-M-performance}

    \centering
    \includegraphics[width=\linewidth]{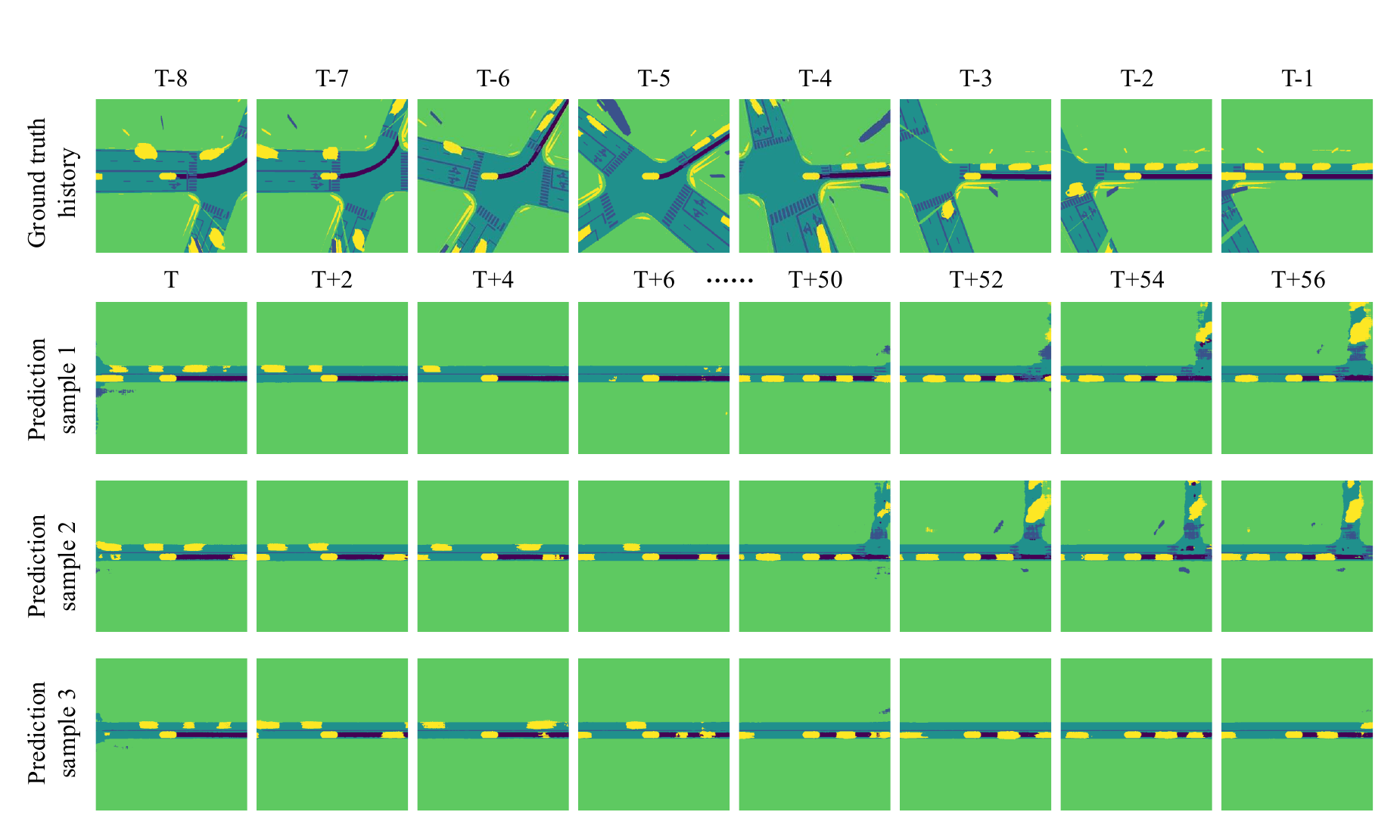}
    \caption{A visualization of the M model's estimated state latent feature obtained by autoregressive rollout.}
    \label{figure: M-imagination}
\end{figure*}

\subsection{Performance on Scenarios from CARLA Leaderboard 2.0}

CARLA Leaderboard 2.0 features significantly higher traffic density and introduces 38 types of challenging interactive traffic scenarios that may be triggered during routes. The average route length in Leaderboard 2.0 exceeds 10 km. Due to computational limitations, running a full test on the original Leaderboard 2.0 takes over a month. As a result, we opted to focus on a subset of routes, each spanning 400m. These routes were selected around special events, with 100m before the event trigger and 300m to ensure proper event activation. Additionally, random weather configurations were applied. Testing on the split routes emphasizes our method’s robustness under varying weather and lighting conditions, with the ability to handle highly interactive traffic scenarios.

In Table \ref{table: performance-scenarios-L20}, we present the performance of RAMBLE on the split routes that highlight specific dynamic traffic scenarios. The model achieves a satisfactory route completion rate across all scenarios, demonstrating that the V and M models effectively handle varying weather and lighting conditions, while the C model is adept at managing diverse traffic situations.

\subsection{Validation of V and M Models}

To assess the effectiveness of the V model in compressing raw sensor data into low-dimensional latent features while preserving crucial information, we visualize the output $z_{T+k}$ decoded by the V model. Additionally, to verify that the M model effectively captures dynamic information, we visualize the predicted $h_{T+k}$, which is based on sensor data from $z_{T+k-8}$ to $z_{T+k-1}$. The visualization results, shown in Figure \ref{figure: V-M-performance}, confirm that both the V and M models successfully capture the essential spatial and temporal information required for the C model to make accurate decisions.

Modeling environmental dynamics is a critical component of model-based RL. To evaluate the M model’s ability to perform multi-step predictions (i.e., ``imagination''), we conduct an autoregressive rollout experiment that forecasts $\{T+k | k=0, 1, \dots, 7\}$ steps. The input to the M model for predicting $\hat{z}_{T+k}$ consists of the sequence $\{z_{T+k-8}, z_{T+k-7}, \dots, \hat{z}_{T+k-2}, \hat{z}_{T+k-1}\}$, along with the action $a_{T+k-1}$. The results of this experiment, visualized using the V model’s VAE decoder, are shown in Figure \ref{figure: M-imagination}. Based on the history features from $T-8$ to $T-1$, the prediction of future $T$ to $T+56$ steps (corresponding to 14s) shows a reasonable imagination of upcoming events.

The V and M models collectively show strong performance in capturing both spatial and temporal features, allowing the C model to make informed decisions. Additionally, the M model’s ability to perform accurate multi-step predictions supports its potential for future event forecasting, making it a promising approach for modeling dynamic environments in reinforcement learning.

\subsection{Ablation Study for the Training Recipe}

\begin{table*}[htb]
    \caption{Ablation study for the training recipe}
    \label{table: ablation-study}
    \centering
    \setlength{\tabcolsep}{4pt}
    \begin{tabular}{@{}>{\raggedright\arraybackslash}p{3.9cm}*{12}{S[table-format=2.2]}@{}}
    \toprule[2pt]
    \textbf{Scenario} &
    \multicolumn{4}{c}{\textbf{RC (\%)}} &
    \multicolumn{4}{c}{\textbf{IP}} &
    \multicolumn{4}{c}{\textbf{DS (\%)}} \\
    \cmidrule(lr){2-5} \cmidrule(lr){6-9} \cmidrule(lr){10-13}
    & {w/o pre.} & {w/ pre.} & {w/ pre.} & {All}
    & {w/o pre.} & {w/ pre.} & {w/ pre.} & {All}
    & {w/o pre.} & {w/ pre.} & {w/ pre.} & {All} \\
    & {+IL} & {+IL} & {+RL} &
    & {+IL} & {+IL} & {+RL} &
    & {+IL} & {+IL} & {+RL} & \\ \midrule
    NonSignalizedJunctionLeftTurn
    & 90.56 & 90.79 & 92.39 & \textbf{97.83}
    & 0.52 & \textbf{0.54} & 0.24 & 0.50
    & 46.21 & \textbf{48.95} & 22.27 & 48.81 \\
    NonSignalizedJunctionRightTurn
    & 91.46 & 93.39 & \textbf{97.75} & 97.37
    & 0.58 & \textbf{0.61} & 0.15 & 0.56
    & 53.21 & \textbf{55.91} & 15.28 & 54.92 \\
    SignalizedJunctionLeftTurn
    & 91.22 & 92.23 & \textbf{97.35} & 94.56
    & 0.52 & 0.53 & 0.27 & \textbf{0.56}
    & 50.24 & 50.77 & 26.12 & \textbf{53.90}\\
    SignalizedJunctionRightTurn
    & 98.43 & 94.46 & 97.34 & \textbf{99.51}
    & 0.61 & \textbf{0.65} & 0.19 & 0.63
    & 60.71 & 61.77 & 18.33 & \textbf{63.01} \\
    OppositeVehicleRunningRedLight
    & 90.26 & 97.01 & 96.17 & \textbf{98.02}
    & 0.48 & 0.74 & 0.29 & \textbf{0.84}
    & 44.98 & 73.50 & 28.69 & \textbf{83.36}\\
    OppositeVehicleTakingPriority
    & 92.35 & 94.20 & 95.71 & \textbf{97.44}
    & 0.44 & 0.74 & 0.31 & \textbf{0.79}
    & 42.55 & 71.93 & 29.73 & \textbf{77.15}\\
    \bottomrule[2pt]
    \end{tabular}
    \vspace{0.1cm}\\
    \scriptsize
    \textit{Notes}:
    ``w/o pre. +IL'': without pre-training of V and M models, only IL, no RL; ``w/ pre. +IL'': with pre-training of V and M models, only IL, no RL;\\
    ``w/ pre. +IL'': with pre-training of V and M models, no IL, only RL; ``All'': with pre-training of V and M models, IL first, RL finetuning.
\end{table*}

\begin{figure*}[htb]
  \centering
  \captionsetup[subfigure]{labelformat=empty}
  \subfloat[(a) Route completion rate vs. Iteration step]{\includegraphics[width=0.48\textwidth]{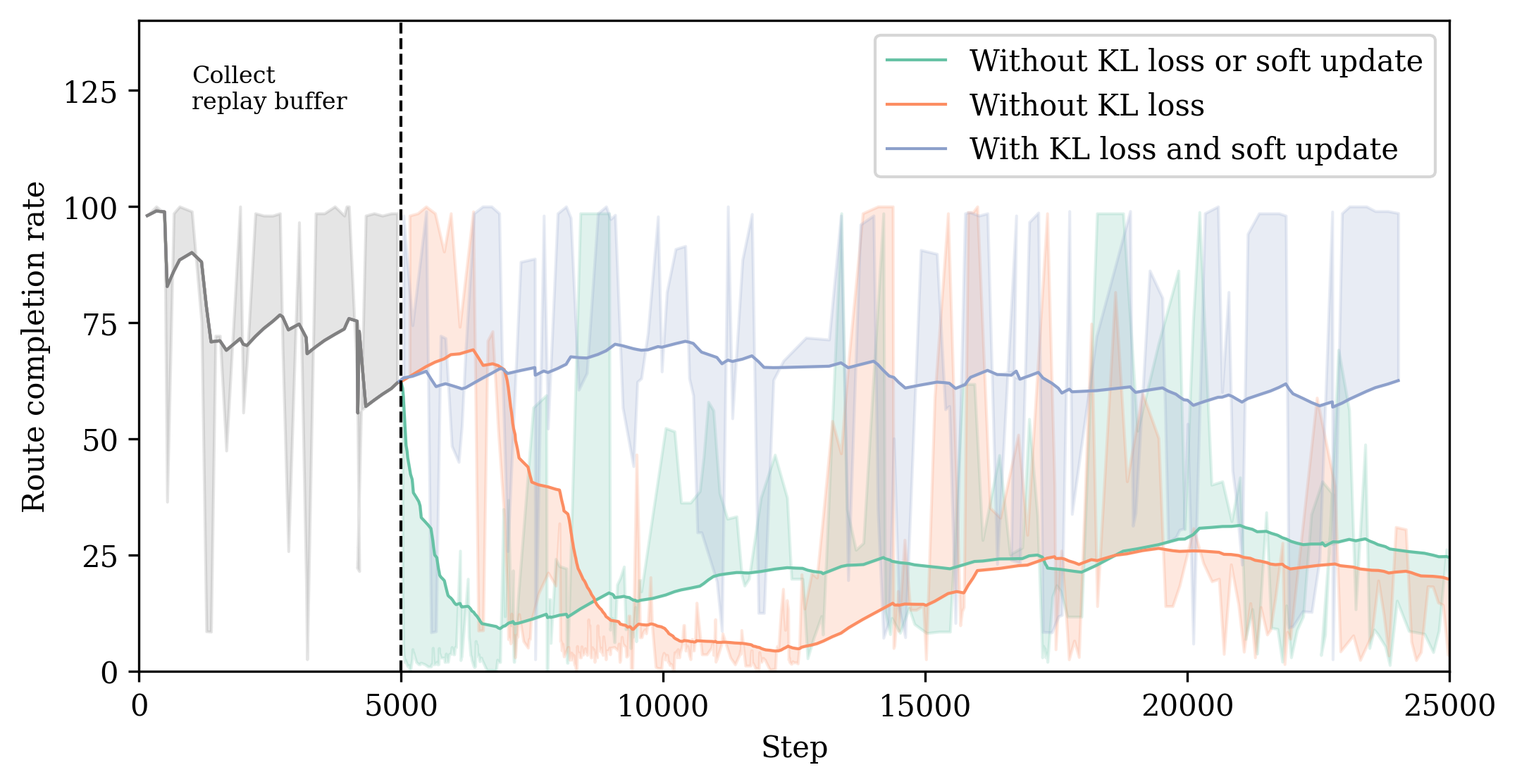}}
  \subfloat[(b) Step reward vs. Iteration step]{\includegraphics[width=0.48\textwidth]{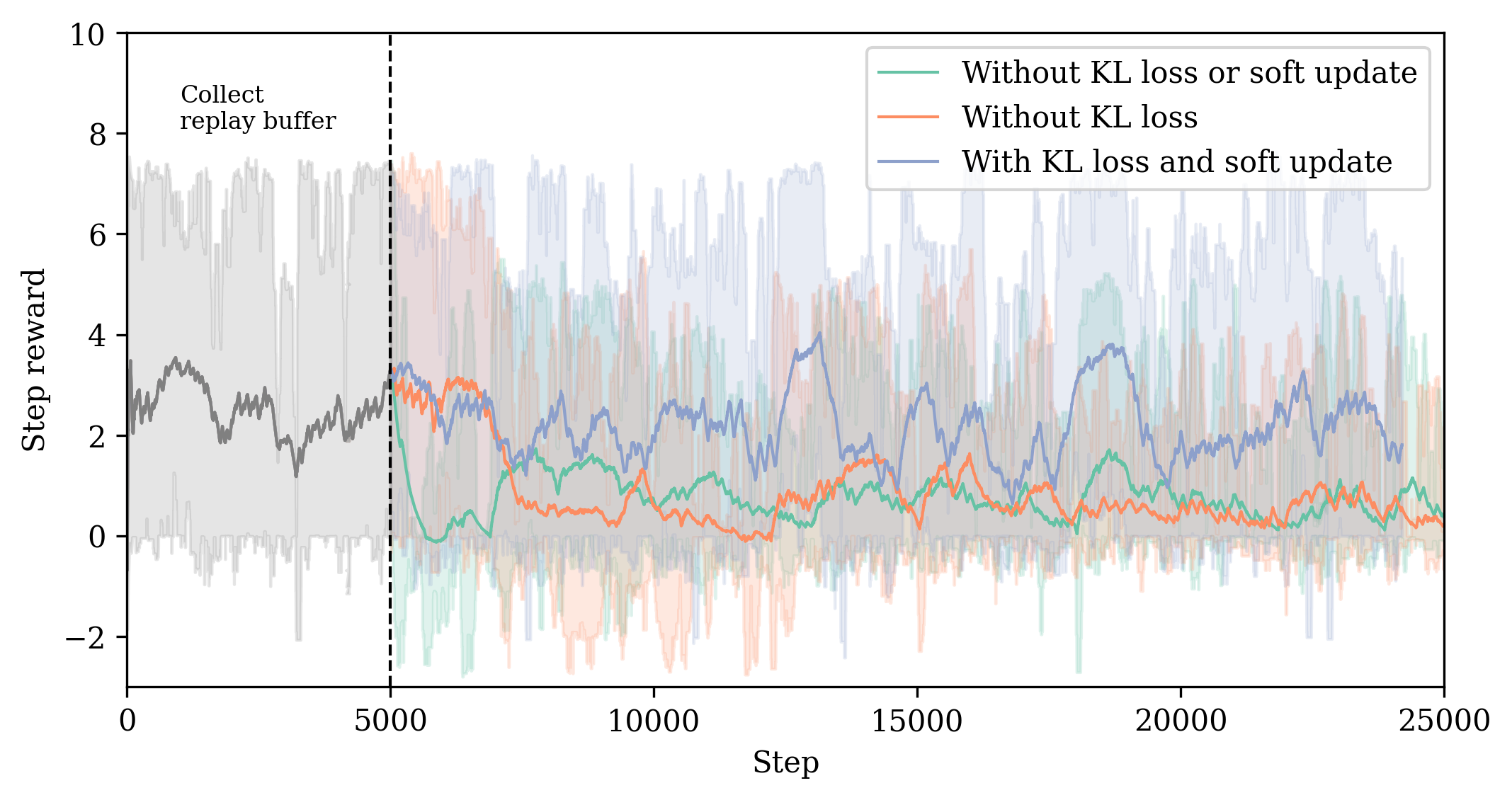}}
  \caption{The C model's performance when it has been pretrained under the IL paradigm and then transitioned to the RL paradigm. The line plots are time-weighted EMAs of the data points with a window size of 50. The shadowed areas depict the true value of the route completion rate and the step reward.}
  \label{figure: ablation-study}
\end{figure*}


Online joint training of the V, M, and C models is feasible once the V and M models have been pre-trained to a sufficient performance level. The convergence speed of the RL agent is heavily constrained by the environment's iteration speed. Even for relatively simple tasks, policy stabilization often requires over 1M steps. However, in the case of the CARLA simulator, this process can take approximately one week. In contrast, IL enables the policy network to achieve convergence within 5-6 hours, significantly accelerating training compared to pure RL.

Table \ref{table: ablation-study} presents an ablation study examining the impacts of omitting pre-training or removing either imitation (IL) or exploration (RL) across various challenging traffic scenarios. Our results show that when the V and M models are trained without pre-training, the driving policy exhibits: (1) significantly prolonged convergence time, (2) reduced route completion rates, and (3) degraded driving scores. These findings underscore the critical role of proper initialization through pre-training for achieving optimal policy performance. While imitation alone proves insufficient for completing certain routes, pure exploration, though achieving higher route completion rates, often produces driving behaviors that significantly diverge from human-like patterns. These findings highlight the critical need for an approach to effectively integrate imitation with exploration.

In Figure \ref{figure: ablation-study}, we validate our method in transferring a model from IL to RL. Due to the nature of decision-making tasks, the actual values of quantitative indicators tend to be unstable and chaotic over long time sequences. To provide a clearer presentation, we display the line plots as the time-weighted Exponential Moving Average of route completion status and step reward, while the shaded areas in different colors represent their actual values, offering a better view of the trends over time. As observed, the KL loss in Equation \ref{equation: rl-with-kl} and the soft update mechanism applied to the anchor network effectively prevent the prior knowledge learned during the IL stage from being overwritten by the highly random exploration that typically occurs at the beginning of RL training, bridging a smooth transition between two different learning paradigms.

\section{Conclusion}

In this paper, we introduce RAMBLE, a novel end-to-end, model-based RL driving model. RAMBLE enhances the RSSM world model by integrating an asymmetric VAE structure to efficiently process high-dimensional, multi-modal sensor data, while utilizing a transformer architecture to capture temporal features. The incorporation of IL pre-training accelerates policy convergence, and the exploration mechanism inherent in RL enables RAMBLE to navigate dynamic and highly interactive traffic scenarios effectively.

RAMBLE achieves SOTA route completion rates on the CARLA Leaderboard 1.0, outperforming leading IL-based approaches. Furthermore, it demonstrates consistently high completion rates across all 38 individual scenarios in CARLA Leaderboard 2.0, underscoring its robustness and ability to generalize across varying weather and lighting conditions, traffic densities, and road types. The results validate the potential of RL in addressing complex traffic scenarios and highlight the critical role of integrating imitation learning and exploration in driving decision-making.

\section{Future Work}

Due to time and computational constraints, we leave the evaluation of our method on the original routes of Leaderboard 2.0 for future work. Additionally, we aim to extend the results of SOTA IL methods from Leaderboard 1.0 to Leaderboard 2.0, establishing a comprehensive benchmark for comparing RL-based and IL-based driving approaches.

We also plan to conduct real-world experiments to validate our method's generalization across diverse driving scenarios. However, significant challenges remain in transferring from simulation to reality, including the need to design and implement effective safety constraints.

\bibliographystyle{IEEEtran}
\bibliography{main}

\begin{IEEEbiography}[{\includegraphics[width=1in,height=1.25in,clip,keepaspectratio]{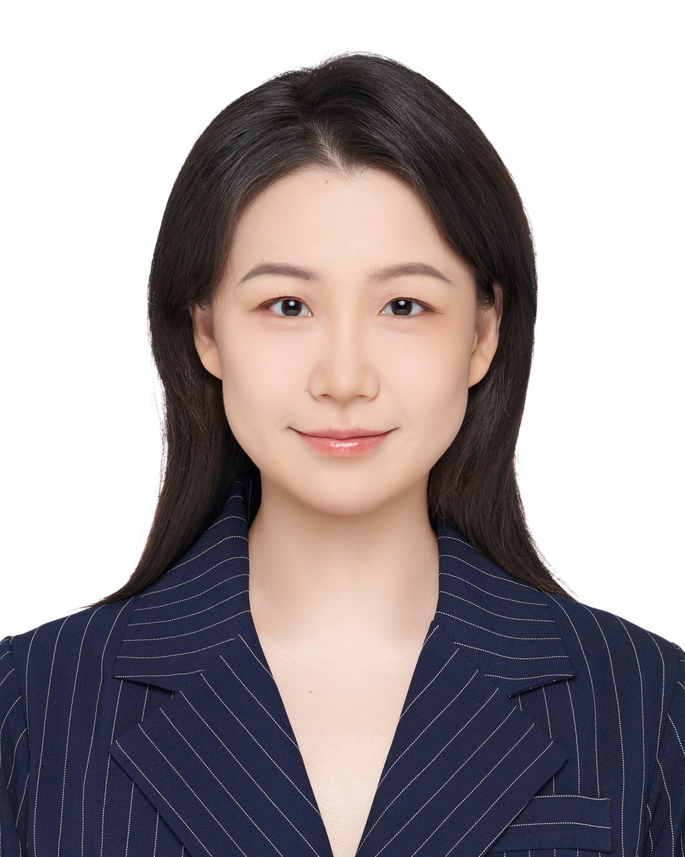}}]{Yueyuan LI}
    received a Bachelor's degree in Electrical and Computer Engineering from the University of Michigan-Shanghai Jiao Tong University Joint Institute, Shanghai, China, in 2020. She is pursuing a Ph.D. degree in Control Science and Engineering from Shanghai Jiao Tong University.

    Her main fields of interest are the security of the autonomous driving system and driving decision-making. Her research activities include reinforcement learning, behavior modeling, and simulation.
\vspace{-20pt}
\end{IEEEbiography}

\begin{IEEEbiography}[{\includegraphics[width=1in,height=1.25in,clip,keepaspectratio]{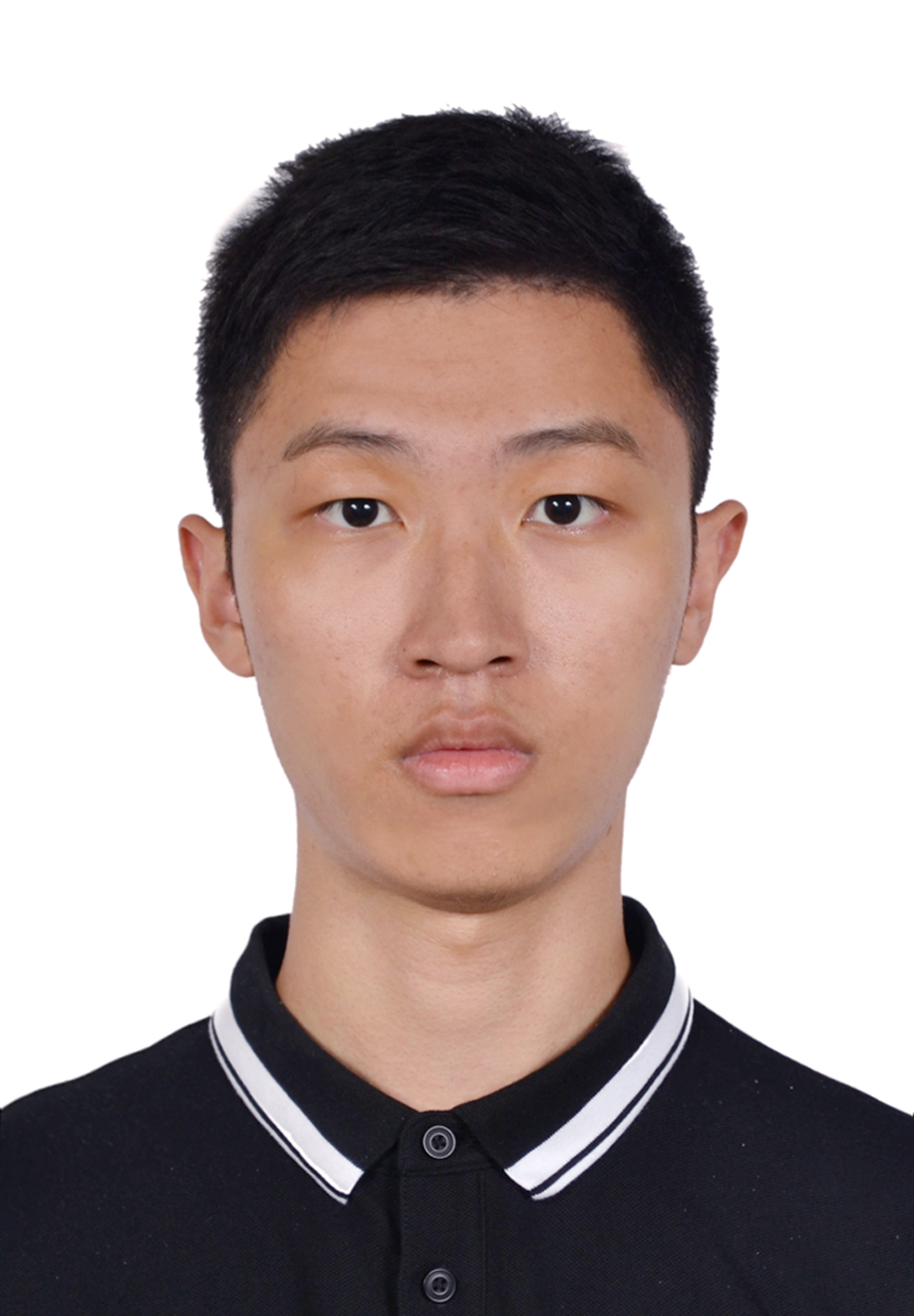}}]{Mingyang JIANG}
    received a Bachelor's degree in engineering from Shanghai Jiao Tong University, Shanghai, China, in 2023. He is working towards a Master's degree in Control Science and Engineering from Shanghai Jiao Tong University. His main research interests are end-to-end planning, driving decision-making, and reinforcement learning for autonomous vehicles.
\vspace{-20pt}
\end{IEEEbiography}

\begin{IEEEbiography}[{\includegraphics[width=1in,height=1.25in,clip,keepaspectratio]{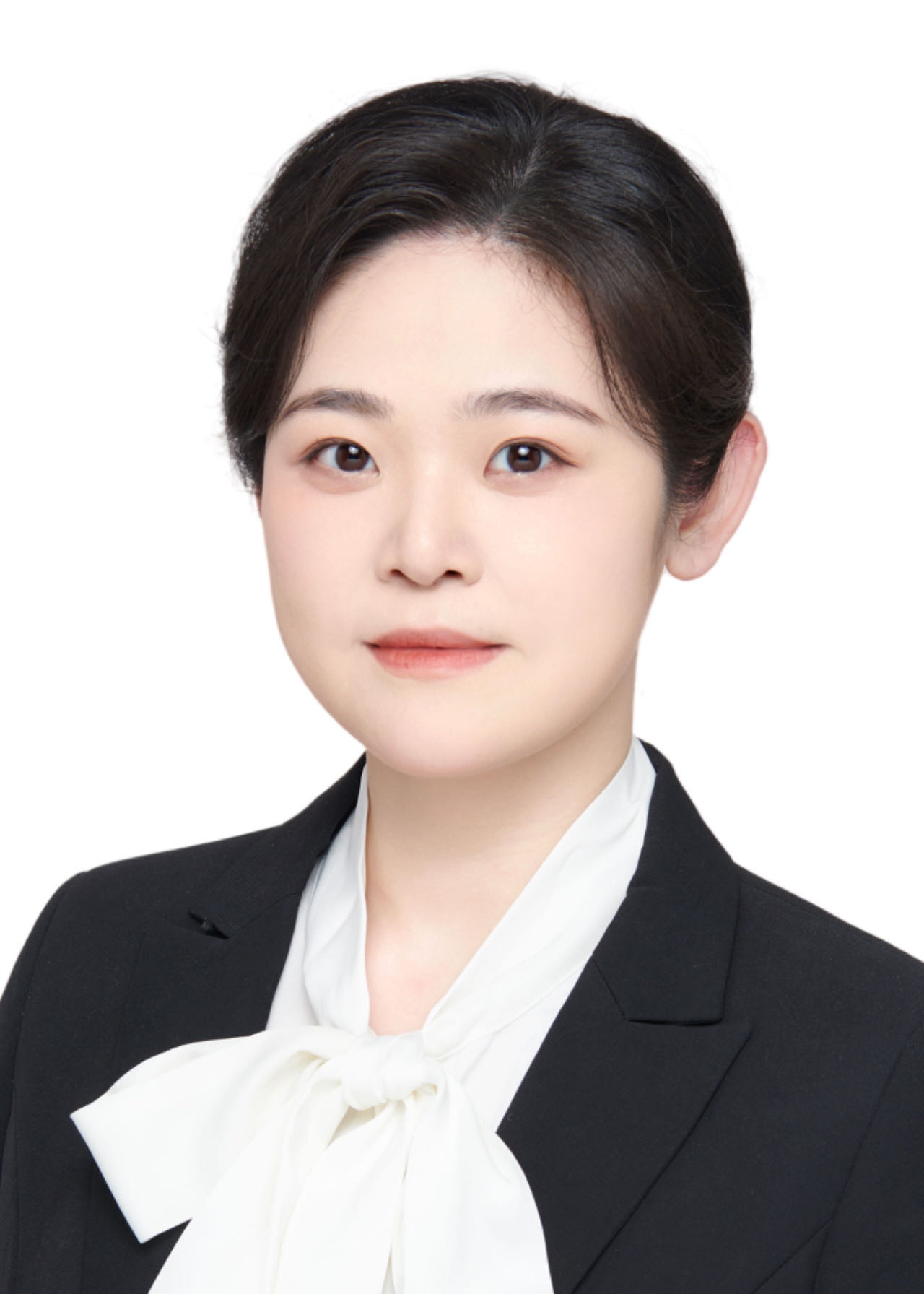}}]{Songan ZHANG}
    received B.S. and M.S. degrees in automotive engineering from Tsinghua University in 2013 and 2016, respectively. Then, she went to the University of Michigan, Ann Arbor, and got a Ph.D. in mechanical engineering in 2021. After graduation, she worked as a research scientist on the Robotics Research Team at Ford Motor Company. Presently, she is an assistant professor at the Global Institute of Future Technology (GIFT) at Shanghai Jiao Tong University. Her research interests include accelerated evaluation of autonomous vehicles, model-based reinforcement learning, and meta-reinforcement learning for autonomous vehicle decision-making.
\vspace{-20pt}
\end{IEEEbiography}

\begin{IEEEbiography}[{\includegraphics[width=1in,height=1.25in,clip,keepaspectratio]{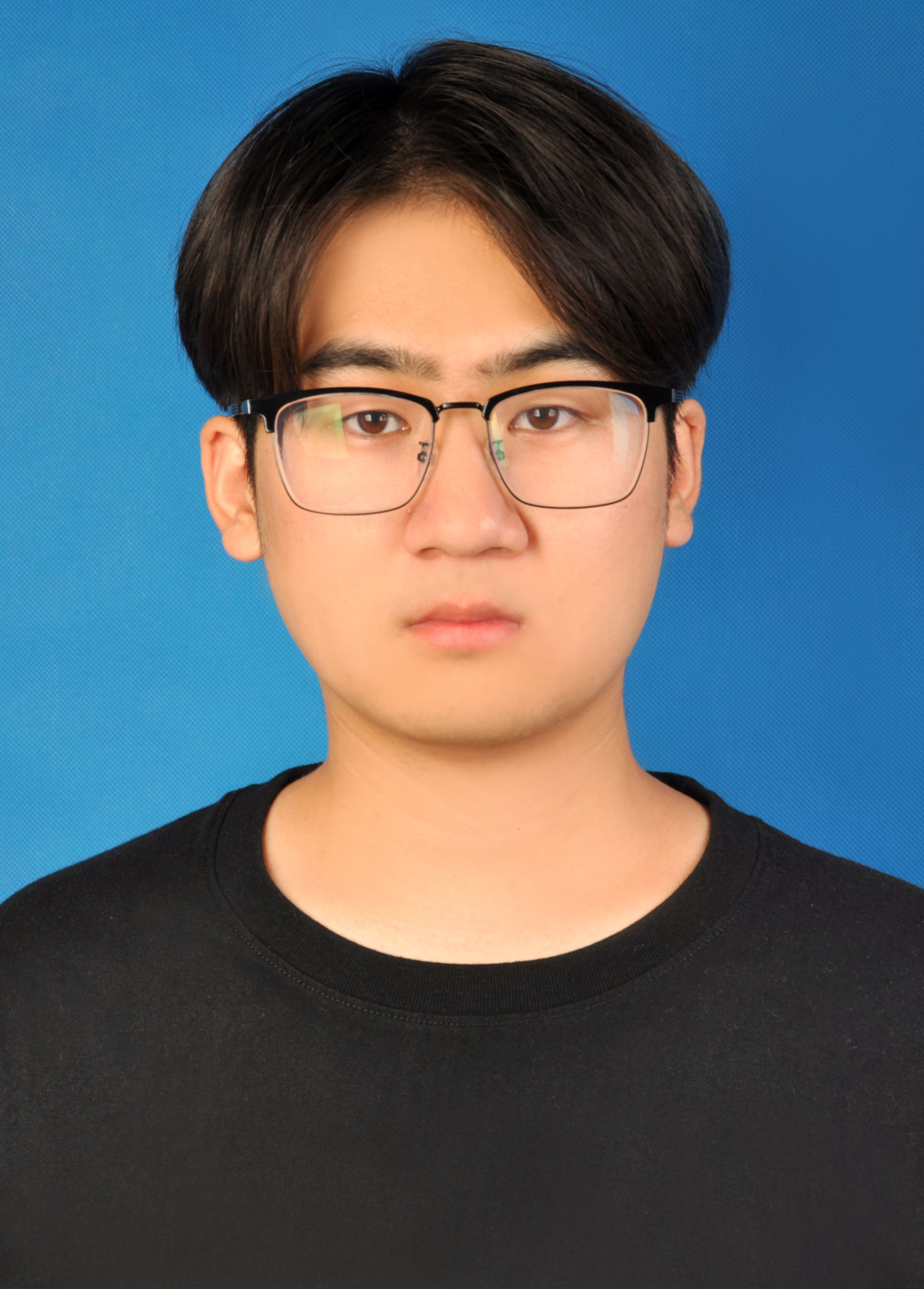}}]{Jiale ZHANG}
    received a Bachelor's degree in Flight Vehicle Control and Information Engineering from Northwestern Polytechnical University, Xi’an, China, in 2023. He is currently pursuing a Ph.D. degree in Control Science and Engineering at Shanghai Jiao Tong University. His main research interests include end-to-end autonomous driving, decision-making algorithms, and reinforcement learning for intelligent vehicles.
\vspace{-20pt}
\end{IEEEbiography}

\begin{IEEEbiography}[{\includegraphics[width=1in,height=1.25in,clip,keepaspectratio]{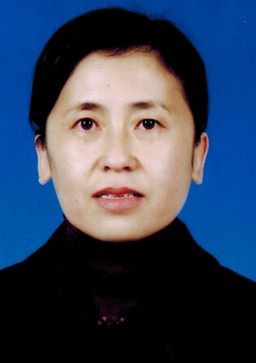}}]{Chunxiang WANG}
    received a Ph.D. degree in Mechanical Engineering from Harbin Institute of Technology, China, in 1999. She is currently an associate professor in the Department of Automation at Shanghai Jiao Tong University, Shanghai, China.

    She has been working in the field of intelligent vehicles for more than ten years and has participated in several related research projects, such as the European CyberC3 project, the ITER transfer cask project, etc. Her research interests include autonomous driving, assisted driving, and mobile robots.
\vspace{-20pt}
\end{IEEEbiography}

\begin{IEEEbiography}[{\includegraphics[width=1in,height=1.25in,clip,keepaspectratio]{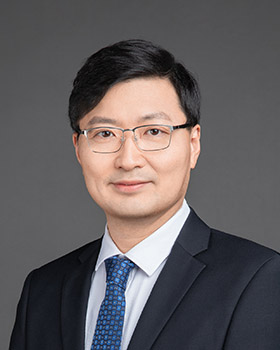}}]{Ming YANG}
    received his Master’s and Ph.D. degrees from Tsinghua University, Beijing, China, in 1999 and 2003, respectively. Presently, he holds the position of Distinguished Professor at Shanghai Jiao Tong University, also serving as the Director of the Innovation Center of Intelligent Connected Vehicles. Dr. Yang has been engaged in the research of intelligent vehicles for more than 25 years.
\vspace{-20pt}
\end{IEEEbiography}






\end{document}